\documentclass[letterpaper, 10 pt, conference]{ieeeconf}  

\IEEEoverridecommandlockouts                              

\usepackage{graphics} 
\usepackage{epsfig} 
\usepackage{amsmath} 
\usepackage{amssymb}  
\usepackage[caption=false,font=footnotesize]{subfig}
\usepackage{multirow}
\usepackage{url}
\usepackage[hidelinks]{hyperref}
\usepackage{csquotes}
\usepackage{comment}

\title{
Multi-Granularity Language-Guided Imitation Learning via Instruction Decomposition 
}

\author{Yi-Pei Chiu and Wei-Ta Chu
\thanks{The authors are with National Cheng Kung University, Taiwan
        {\tt\small p76134684@gs.ncku.edu.tw, wtchu@gs.ncku.edu.tw}}%
}

\begin{document}

\maketitle
\thispagestyle{empty}
\pagestyle{empty}

\begin{abstract}
Using language instructions as conditions to guide robot policy learning has recently become an important research domain. However, existing language-guided policy learning methods typically use an overall task description to guide the entire demonstration trajectory. For manipulation tasks involving multiple execution stages, these methods assign the same language description to different subtasks, making it difficult to distinguish the behaviors required at different stages. In this work, we propose a multi-granularity language guidance method based on instruction decomposition. The proposed method decomposes an overall task description into more fine-grained, concrete subtask-level language instructions, thereby enhancing learning efficiency and improving performance. We evaluate the proposed method in the setting of multi-task imitation learning and validate its effectiveness.
\end{abstract}

\section{INTRODUCTION}
A task described by natural language is often not a single atomic action, but a composition of multiple intermediate steps that must be executed in a coherent order. Humans can naturally infer such procedural structure from high-level instructions, decomposing an abstract goal into a sequence of motion primitives. However, this remains challenging for robots. They must not only ground the language instruction in the physical environment, but also determine how the specified goal can be achieved through temporally extended behaviors. Recent studies in language-guided robotic policy learning have adopted natural language as a task condition~\cite{lynch2020language}\cite{Distill-Down}\cite{MDT}. Nevertheless, most existing approaches condition the policy on a single language goal and directly learn the corresponding behavior, without explicitly modeling the intermediate steps. 

\begin{figure}[t]
  \centering
  \includegraphics[width=\linewidth]{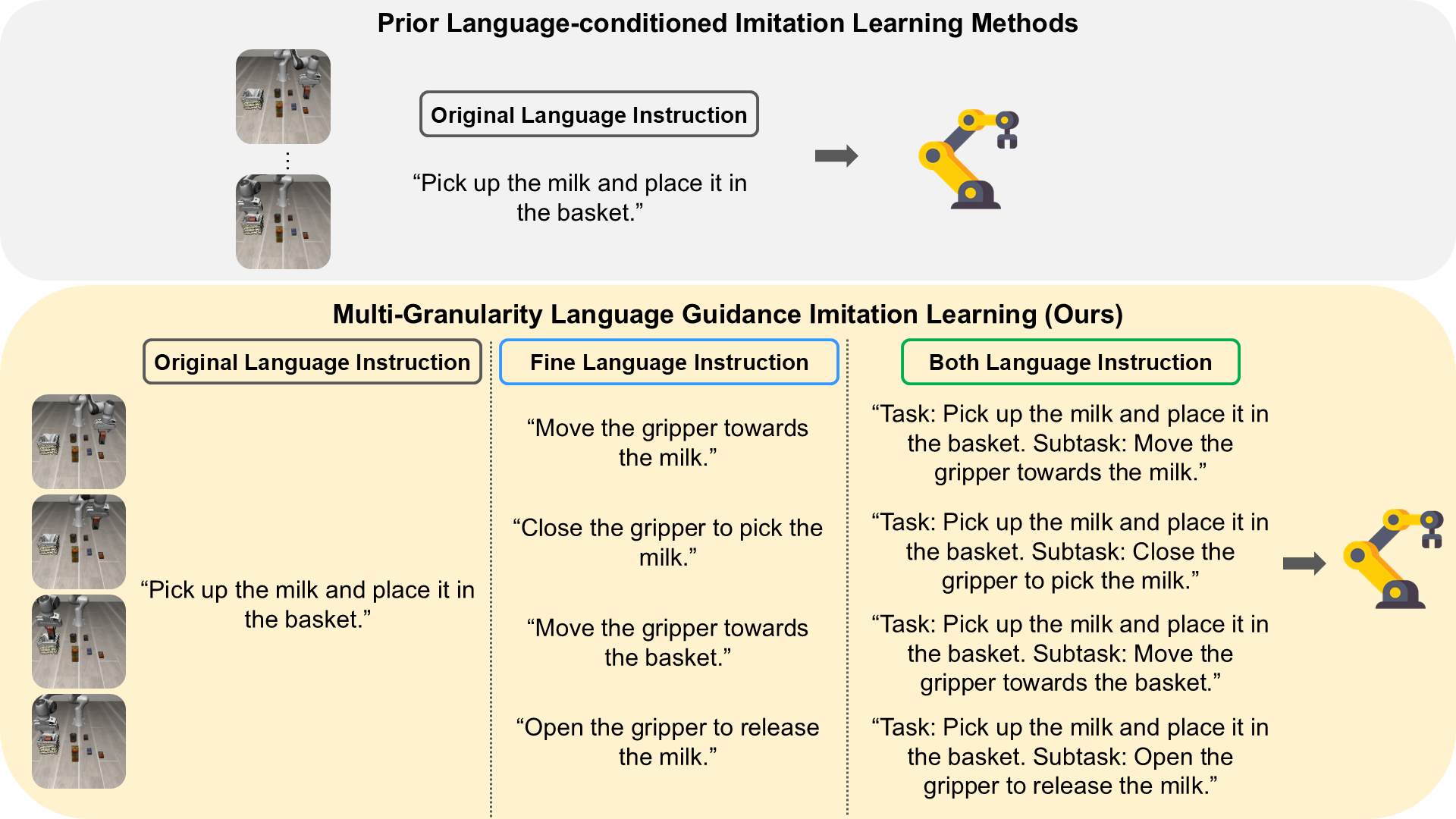}
  \caption{Comparison of language guidance strategies in imitation learning. Prior methods use a single overall instruction for the entire trajectory, while our method introduces multi-granularity language guidance by incorporating task-level instructions, fine-grained subtask-level instructions, and their combination during policy learning.}
  \label{fig:comparison}
\end{figure}

As shown in Figure~\ref{fig:comparison}, prior language-guided imitation learning methods typically rely on a single overall instruction to condition the entire trajectory. However, such coarse language supervision often fails to distinguish the different sub-goals and action requirements involved in a multi-step task. For example, a high-level instruction such as \enquote{Pick up the object and place it into the container} does not provide explicit guidance for each intermediate stage, such as moving the gripper above the object, grasping the object, lifting it, moving it above the container, and opening the gripper to release it. As a result, the policy must learn substantially different behaviors under the same overall language condition. This makes it difficult to associate each stage of the trajectory with the appropriate action, and may lead to ambiguous or misaligned behaviors, especially in long-horizon or multi-stage manipulation tasks.

Motivated by these observations, we propose \textbf{MuGIL}, a \textbf{Mu}lti-\textbf{G}ranularity language guidance framework for \textbf{I}mitation \textbf{L}earning. Instead of conditioning the policy solely on a single task-level instruction, MuGIL leverages language guidance at multiple granularities during training. Specifically, the policy is trained to capture the global task objective from the overall instruction while also learning fine-grained action guidance from subtask-level descriptions. 

The rest of this paper is organized as follows. Sec.~\ref{chap:related} presents related works on language-guided policy learning in several aspects. Sec.~\ref{chap:method} provides details of the proposed MuGIL framework. Sec.~\ref{chap:exp} describes the evaluation results and ablation studies, followed by the concluding remarks in Sec.~\ref{chap:conclusion}. 

\section{RELATED WORK}
\label{chap:related}
In recent years, diffusion models have attracted growing attention in robotic learning. Diffusion-based methods learn to generate action sequences by iteratively denoising Gaussian-corrupted actions conditioned on observations~\cite{GaussianNoise}\cite{song2020score}. Meanwhile, language has become an increasingly important representation of task goals in robotic manipulation. In imitation learning, many approaches encode natural language task descriptions into embeddings using pretrained language models, which are then provided as conditional inputs to policies trained on multi-task datasets. 

Distill-Down~\cite{Distill-Down} extends a single-task diffusion policy~\cite{DP} to multi-task policy learning. PlayFusion~\cite{Playfusion} applies diffusion-based policy learning to unstructured and suboptimal play data, 
where language annotations are used to specify goal-directed skills. MDT~\cite{MDT} further explores diffusion policies with multimodal goals by incorporating both language and target-image goals during training. Together, these works demonstrate the effectiveness of language-guided diffusion-based policies in robotic manipulation. However, all these methods take a single overall instruction to guide the entire trajectory. 

Some studies \cite{PALO}\cite{CLAP}\cite{RACER}\cite{Steer} show that decomposing complex, high-level instructions into low-level subgoals helps robots complete sophisticated tasks more reliably. PALO~\cite{PALO} employs vision-language models to translate high-level task descriptions into reusable subtasks, enabling rapid adaptation with minimal supervision. CLAP~\cite{CLAP} decomposes high-level task instructions into step-wise instructions and uses them to guide language-aligned 3D keypoint prediction. These works show that decomposing high-level instructions into more low-level subgoals can improve policy adaptation and generalization. 
 
RACER~\cite{RACER} augments expert demonstrations with failure recovery trajectories and fine-grained language annotations, allowing the policy to learn how to recover from execution failures. STEER~\cite{Steer} relabels existing robot demonstrations with dense natural language commands that describe modular manipulation skills, enabling the learned policy to be controlled not only by what task to perform but also by how the behavior should be executed. These works show that richer language can provide sufficient details beyond just simple language instructions. 

Like these works, our method also provides richer language guidance to policy learning. Previous works often use decomposed or richer language for policy adaptation, 3D keypoint prediction, failure recovery, or reasoning-based generalization. On the other hand, we focus on the effect of multi-granularity language guidance in imitation learning. 
Specifically, we jointly incorporate task-level instructions, fine-grained subtask-level instructions, and their combination during training. Rather than treating fine-grained language as a replacement for task-level instructions, we study how these different levels of language guidance can complement each other during policy learning.

\begin{figure}
  \centering
  \includegraphics[width=\linewidth]{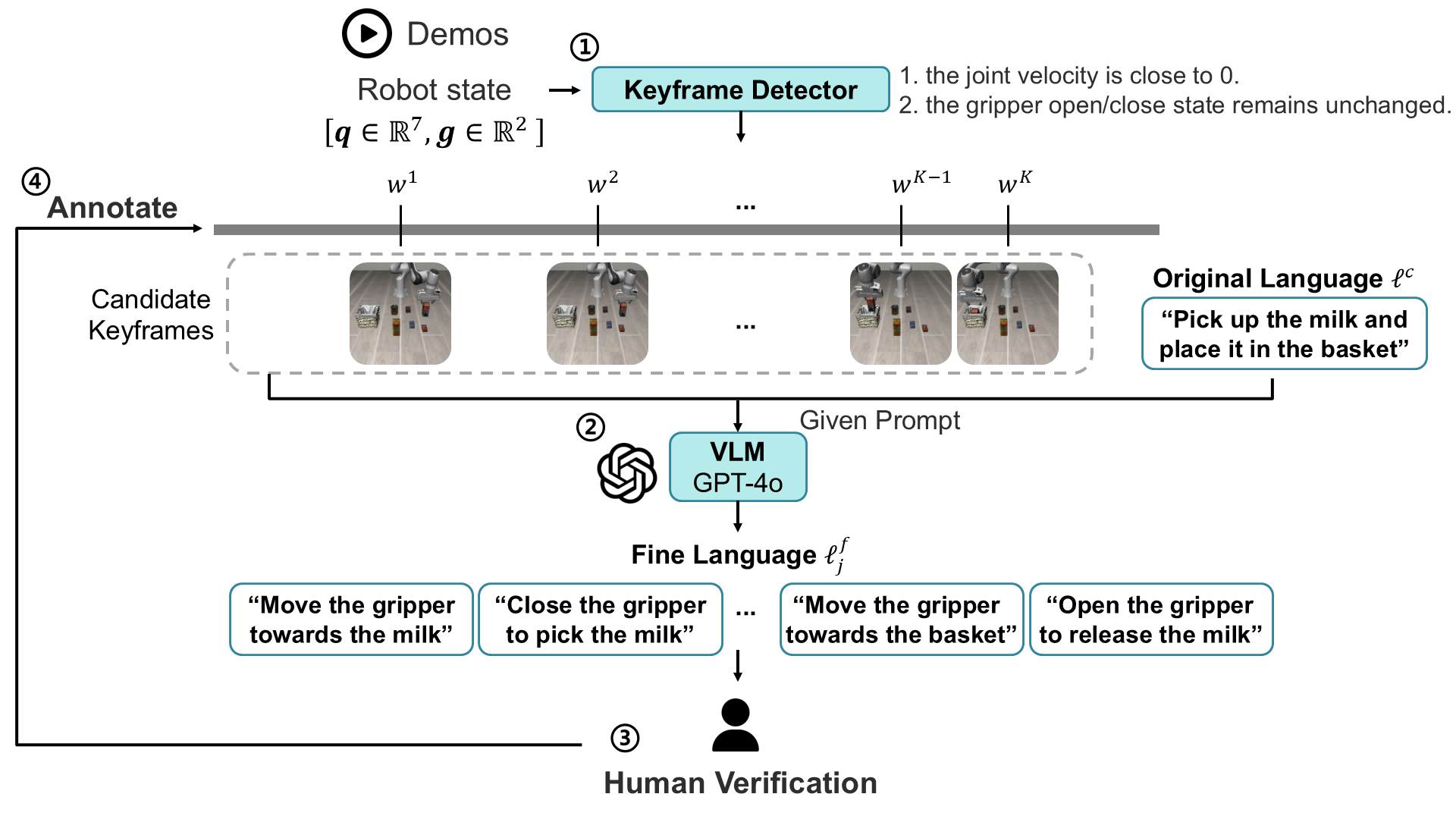}
  \caption{Illustration of the fine-grained language annotation process. 
  Step 1: Detect keyframes based on robot states. 
  Step 2: Generate fine language instructions by a VLM. 
  Step 3: Human verify to reduce annotation mess. 
  Step 4: Assigning fine language instructions to corresponding subtask segments.}
  \label{fig:annotation}
\end{figure}


\begin{figure*}
  \centering
  \includegraphics[width=12cm]{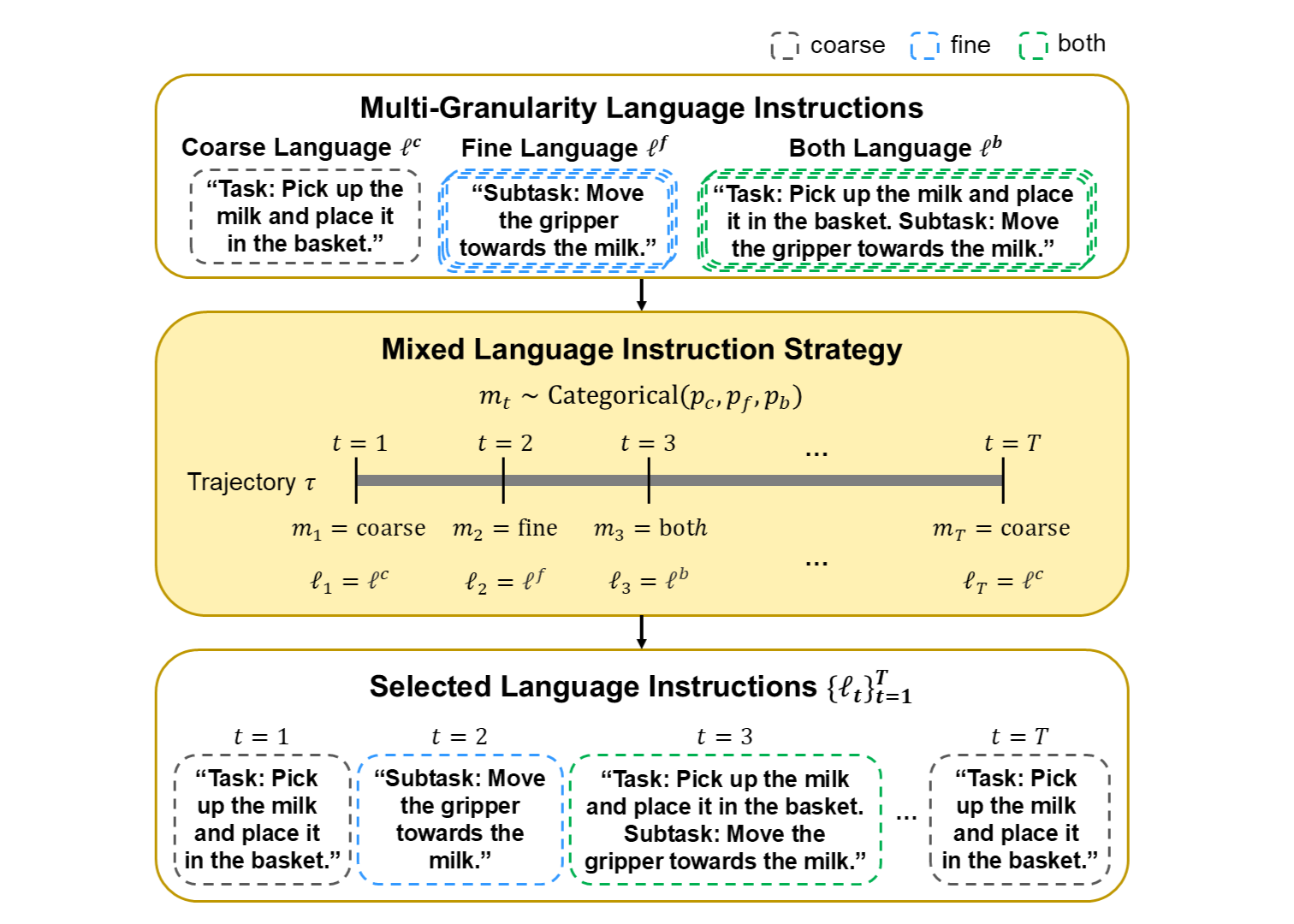}
  \caption{Illustration of the mixed language instruction strategy.
  For each timestep $t$ in a trajectory $\tau$, a language mode $m$ is sampled from coarse $\ell^c$, fine $\ell^f$, and both $\ell^b$ language instructions.}
  \label{fig:mixed_strategy}
\end{figure*}

\section{METHOD}
\label{chap:method}

\subsection{Problem Formulation}
\label{sec:problem_formulation}
In language-guided imitation learning, the objective is to train a policy that predicts robot actions from visual observations, conditioned on natural language instructions. Each individual trajectory is represented as $\tau = \left\{(o_t, a_t)\right\}_{t=1}^{T}$, where $o_t$ denotes the observation image at timestep $t$, and $a_t$ denotes the corresponding robot action.

Unlike conventional language-guided imitation learning settings, where a single language instruction $\ell$ is assigned to an entire trajectory, we extend a single language instruction into temporally aligned language instructions $\ell_i$. Each timestep is associated with a corresponding language condition. Specifically, starting from the overall language annotation for each trajectory, we decompose it into multiple finer-grained language instructions. Based on this process, we construct a multi-granularity language-guided dataset $\mathcal{D}=\left\{\left(o_i, \bar{a}_i, \ell_i\right)\right\}_{i=1}^{N}$, where $N$ is the total number of samples collected from all trajectories, and the language condition $\ell_i$ may represent descriptions at different levels, including an original task-level instruction, a finer subtask-level instruction, or their combination. The goal is to learn a language-guided policy $\pi_{\theta}\left(\bar{a}_i \mid o_i, \ell_i\right)$ that predicts a sequence of actions $\bar{a}_i=\left(a_i, a_{i+1}, \ldots, a_{i+h-1}\right)$ of length $h$, conditioned on the current observation $o_i$ and the corresponding language instructions $\ell_i$. Our policy is trained to maximize the log-likelihood of the action sequence given the observation and language goal: 

\begin{equation}
    \mathbb{E}
    \left[
    \sum_{(o_i,\bar{a}_i,\ell_i)\in\mathcal{D}}
    \log \pi_{\theta}\left(\bar{a}_i \mid o_i, \ell_i\right)
    \right].
    \label{eq:il_loss}
\end{equation}

\subsection{Fine-Grained Language Annotation}
\label{sec:fine_grained_language_annotation}
Most existing language-guided datasets~\cite{LIBERO}\cite{Rlbench}\cite{CLIPort} provide a single task-level language instruction for an entire demonstration. The intermediate execution stages are not explicitly provided to guide detailed sub-goals. To provide more detailed language supervision, we decompose the original language annotation for each demonstration into finer-grained subtask-level annotations. Figure~\ref{fig:annotation} illustrates the process.

The LIBERO dataset~\cite{LIBERO} used in this work is collected with a 7-DoF Franka Emika Panda robot. In addition to visual observations and robot actions, the dataset also contains proprioceptive robot states, including the 7-dimensional joint state $\textbf{q}_t \in \mathbb{R}^{7}$ and the 2-dimensional gripper state $\textbf{g}_t \in \mathbb{R}^{2}$. To identify boundaries between subtasks, we first detect keyframes based on the robot's state. Following prior works~\cite{Q-attention}\cite{HDP}, a timestep is selected as a keyframe candidate if (1) the joint velocity is close to 0; and (2) the gripper open/close state remains unchanged. The intuition is that when the robot temporarily slows down and maintains a stable gripper state, it often reaches an intermediate state that may correspond to the beginning or the end of a subtask. After keyframe detection, let $\mathcal{W} = \{ w_1, w_2, \ldots, w_K \}$ denote the set of detected keyframes in a demonstration, where $w_j$ denotes the timestep index of the $j$-th keyframe, and $K$ denotes the total number of detected keyframes in the demonstration. 

Given the detected keyframes $\mathcal{W}$ and the original language instruction $\ell^c$, we use GPT-4o~\cite{GPT4} to generate fine language instructions based on the original language instruction, the keyframe, and a text prompt. The generated fine instruction is expected to describe a local objective of the current subtask, such as \enquote{Move the gripper towards the object} and \enquote{Close the gripper to pick the object}. The fine language instruction for the $j$-th keyframe is denoted as $\ell_{j}^{f}$ in the following. Since automatically generated annotations may not perfect, we manually inspect and correct them. This verification step is important for reducing annotation mess and ensuring that the fine language instructions are semantically consistent with the corresponding robot behaviors. 

Finally, the fine language instructions are assigned back to the timesteps of the original demonstration. For each detected keyframe $w_j$, we define a subtask segment as the temporal interval from the previous keyframe to the current keyframe, i.e., $[w_{j-1}, w_{j}]$. The fine instruction generated for this segment is denoted as $\ell_j$. Then, for every timestep $t$ within this segment, the same fine language instruction is assigned as
\begin{equation}
    \ell(t) = \ell_{j},
    \quad
    t \in [w_{j-1}, w_{j}].
    \label{eq:fine_langauge_instructions}
\end{equation}
In this way, all movements within the same subtask segment are associated with the same fine language instruction.

Through this annotation process, we obtain a multi-granularity language annotated dataset, where each demonstration contains both the original task-level instruction 
and the finer subtask-level instructions. 

\begin{figure*}
  \centering
  \includegraphics[width=12cm]{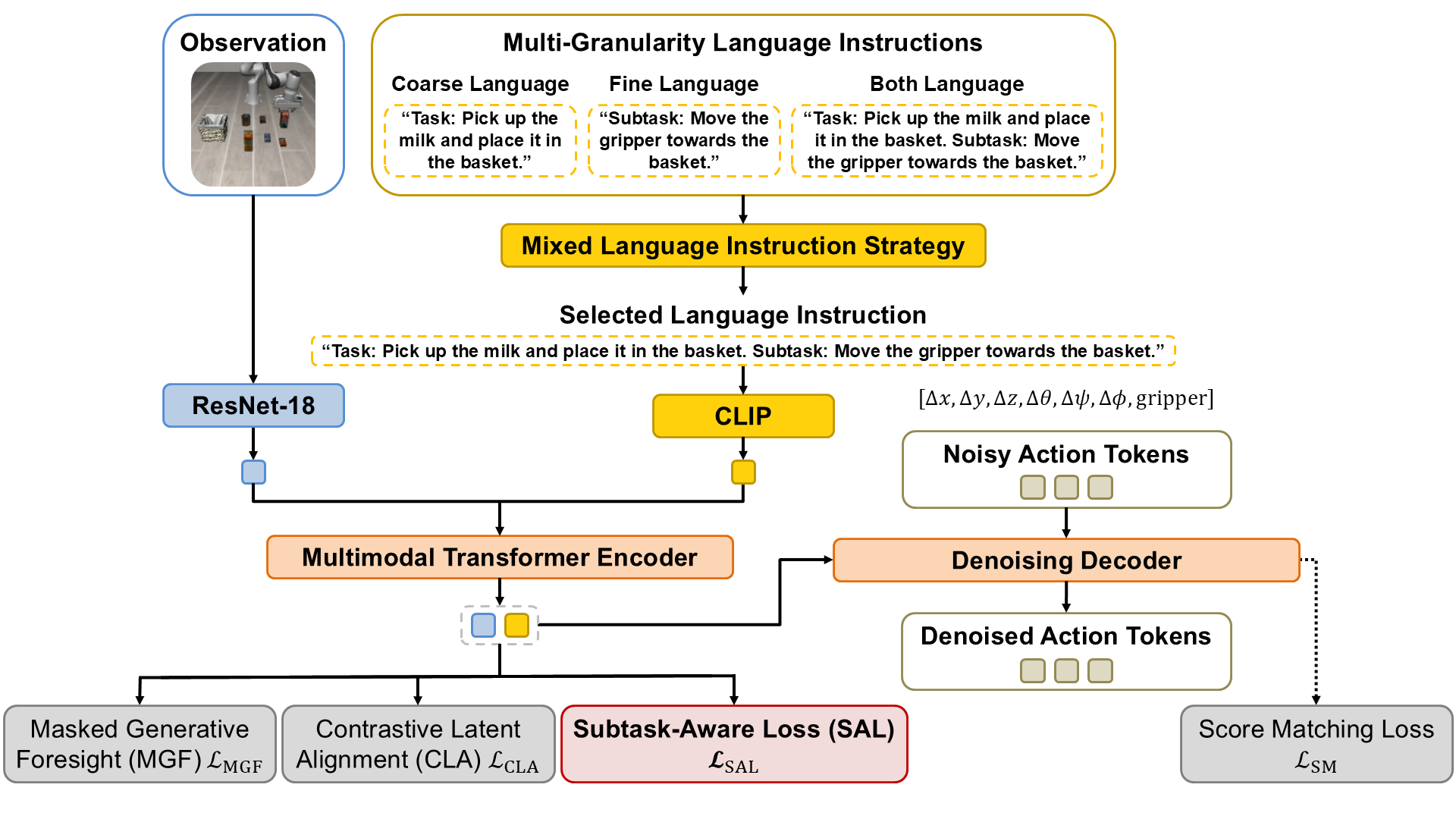}
  \caption{Illustration of the proposed Multi-Granularity Language Guidance for Imitation Learning (MuGIL) method.}
  \label{fig:policy_learning}
\end{figure*}

\subsection{Mixed Language Instruction Strategy}
\label{sec:mixed_language_instruction_strategy}
Now each demonstration is associated with language instructions at two different levels. To train the policy, we consider three language-guided modes: coarse, fine, and both language instructions. The coarse language instruction describes the overall task objective. The fine language instruction describes the current subtask stage and provides more detailed local information. The both setting combines the coarse and fine language instructions, thereby containing both the global task objective and the local subtask context.

We introduce a mixed language instruction strategy that allows the policy to learn from multiple forms of language guidance, as shown in Figure~\ref{fig:mixed_strategy}. For each timestep $t$ in a trajectory $\tau$, we select a language instruction $\ell_t$ according to one of the three language modes. 
The language instruction used for training is defined as 
\begin{equation}
    \ell_t =
    \begin{cases}
    \ell_t^c, & m_t = \mathrm{coarse}, \\
    \ell_t^f, & m_t = \mathrm{fine}, \\
    \ell_t^b, & m_t = \mathrm{both},
    \end{cases}
    \quad
    m_t \in \{\mathrm{coarse}, \mathrm{fine}, \mathrm{both}\}.
    \label{eq:language_modes}
\end{equation}

Since the coarse instruction describes the overall task objective, we set $\ell_t^c = \ell^c$ for all timesteps in the same trajectory. The both language instruction is constructed by combining the coarse and fine language instructions: $\ell_t^b = \mathrm{Concat}(\ell^c, \ell_t^f).$

During training, the language mode is sampled for each timestep according to a categorical distribution:
\begin{equation}
    m_t \sim \mathrm{Categorical}(p_c, p_f, p_b),
    \label{eq:language_mode_distribution}
\end{equation}
where $p_c$, $p_f$, and $p_b$ denote the sampling probabilities for the coarse, fine, and both language modes, respectively. In our experiments, we empirically set the sampling ratio as $p_c : p_f : p_b = 0.6 : 0.1 : 0.3$. We further investigate the influence of different sampling ratios through an ablation study in Section~\ref{sec:sampling_ratio}. 

Given a trajectory with $T$ timesteps, this strategy samples one of the three language modes for each timestep $t$. Different timesteps within the same trajectory may 
be conditioned on language instructions at different levels. This design allows the policy to learn from complementary information provided by different language modes. Therefore, the proposed mixed language instruction strategy enables the policy to benefit from multi-granularity language guidance during policy learning.

\subsection{Language-Guided Policy Learning}
\label{sec:language_conditioned_policy_learning}

\subsubsection{Diffusion-Based Policy Architecture}
\label{sec:policy_architecture}
Figure~\ref{fig:policy_learning} illustrates the language-guided policy learning process. We adopt MDT~\cite{MDT} as the base architecture for language-guided imitation learning. To enable MDT to leverage partially annotated datasets~\cite{CALVIN, LfP}, it is designed to learn from multimodal goals, including language goals and visual goals. In our work, we focus on evaluating the effect of language guidance in policy learning. Therefore, we use language as the primary goal condition, while retaining the visual goal for auxiliary objectives introduced in MDT. 

MDT uses an encoder-decoder transformer~\cite{Attention} architecture for diffusion-based action prediction. Given the current visual observation and the language goal, the encoder converts the conditioning inputs into a set of latent representation tokens. The decoder acts as a diffusion denoiser, which predicts the action sequence by iteratively denoising noisy actions from the encoder outputs. 

For visual observations, MDT uses a ResNet-18 encoder~\cite{DP} to extract observation features, which are represented as an observation token. For language conditioning, the language instruction is encoded by a frozen CLIP text encoder~\cite{CLIP} and represented as a language token. Different from conventional language-guided policies that use a single task-level language instruction for the entire trajectory, our method adopts the mixed language instruction strategy mentioned above. After obtaining the observation and language tokens, the MDT encoder processes these tokens through several self-attention transformer layers and produces latent representations.

The MDT decoder generates the action sequence through a diffusion denoising process. During training, Gaussian noise is added to an action sequence, 
and the decoder is trained to denoise the noisy action sequence. In each decoder layer, cross-attention is used to incorporate the conditioning information 
from the encoder outputs into the denoising process. The current noise level $\sigma_k$ is encoded by a sinusoidal embedding followed by an Multi-Layer Perceptron (MLP) that produce a latent noise token, which is further injected into the transformer decoder blocks.

Following diffusion-based policy learning, MDT generates actions by denoising noisy action sequences. Given the visual observation $o_i$ and the language instruction $\ell_i$, a neural network $D_{\theta}$ is trained to approximate the score function of the diffusion process via Score Matching (SM)~\cite{SM}:
\begin{equation}
    \mathcal{L}_{\mathrm{SM}}
    =
    \mathbb{E}_{\sigma, \bar{a}_i, \epsilon}
    \left[
    \alpha(\sigma_k)
    \left\|
    D_{\theta}
    \left(
    \bar{a}_i + \epsilon,
    o_i,
    \ell_i,
    \sigma_k
    \right)
    -
    \bar{a}_i
    \right\|_2^2
    \right],
    \label{eq:sm_loss}
\end{equation}
where $\epsilon$ denotes the sampled Gaussian noise~\cite{GaussianNoise}, $\sigma_k$ denotes the noise level at diffusion step $k$, and $\alpha(\sigma_k)$ is a weighting function that depends on the noise level. During training, noise is sampled randomly from a noise distribution and added to the ground-truth action sequence. The denoising network $D_{\theta}$ then predicts the denoised actions and is optimized by minimizing the score matching loss.

In addition to the diffusion objective, MDT introduces two auxiliary objectives: Masked Generative Foresight (MGF) and Contrastive Latent Alignment (CLA).
MGF is used to encourage the latent embedding to predict future observation information. Given the current observation $o_i$, MGF reconstructs the image patches of a future observation $o_{i+v}$, where $v$ denotes the foresight distance. Let $(\mathbf{u}_1,\ldots,\mathbf{u}_U)=\operatorname{patch}(o_{i+v})$ denote the $U$ image patches extracted from $o_{i+v}$. The MGF loss is defined as
\begin{equation}
    \mathcal{L}_{\mathrm{MGF}}(o_i)
    =
    \frac{1}{U}
    \sum_{\mathbf{u} \in \operatorname{patch}(o_{i+v})}
    \mathbf{1}_{\mathrm{m}}(\mathbf{u})
    \left(\mathbf{u} - \hat{\mathbf{u}}\right)^2,
    \label{eq:mgf_loss} 
\end{equation}
where $\hat{\mathbf{u}}$ is the reconstructed patch corresponding to $\mathbf{u}$, and the indicator function $\mathbf{1}_{\mathrm{m}}(\mathbf{u})$ is $1$ if $\mathbf{u}$ is masked and $0$ otherwise. 

CLA is an auxiliary objective to align visual and language goal representations using contrastive learning~\cite{ContrastiveLearning}. Given a training batch of size $B$ with multimodal goals, MDT obtains two normalized embeddings, $\mathbf{z}_i^{\mathrm{o}}$ and $\mathbf{z}_i^{\mathrm{l}}$, which correspond to the observation-goal conditioned state embedding and the language-goal conditioned state embedding, respectively. These embeddings are obtained by pooling the MDT latent tokens into a single vector for each goal modality. The CLA loss is computed using a symmetric InfoNCE~\cite{InfoNCE} objective:
\begin{equation}
    \begin{aligned}
    \mathcal{L}_{\mathrm{CLA}}
    = -\frac{1}{2B}\sum_{i=1}^{B}
    \Bigg[
    &\log
    \left(
    \frac{
    \exp\left(\frac{C(\mathbf{z}_i^{\mathrm{o}}, \mathbf{z}_i^{\mathrm{l}})}{v}\right)
    }{
    \sum_{j=1}^{B}
    \exp\left(\frac{C(\mathbf{z}_i^{\mathrm{o}}, \mathbf{z}_j^{\mathrm{l}})}{v}\right)
    }
    \right) \\
    &+
    \log
    \left(
    \frac{
    \exp\left(\frac{C(\mathbf{z}_i^{\mathrm{o}}, \mathbf{z}_i^{\mathrm{l}})}{v}\right)
    }{
    \sum_{j=1}^{B}
    \exp\left(\frac{C(\mathbf{z}_j^{\mathrm{o}}, \mathbf{z}_i^{\mathrm{l}})}{v}\right)
    }
    \right)
    \Bigg].
    \end{aligned}
    \label{eq:cla_loss}
\end{equation}
with temperature parameter $v$. The MDT training objective is a weighted sum of the score matching loss and the auxiliary losses:
\begin{equation}
    \mathcal{L}_{\mathrm{MDT}}
    =
    \mathcal{L}_{\mathrm{SM}}
    +
    \alpha \mathcal{L}_{\mathrm{MGF}}
    +
    \beta \mathcal{L}_{\mathrm{CLA}},
\label{eq:mdt_loss}
\end{equation}
where $\alpha=0.1$ and $\beta=0.1$ in our experiments. 

\subsubsection{Subtask-Aware Loss}
\label{sec:subtask_aware_loss}
In addition to the training objectives described above, we introduce an auxiliary objective called Subtask-Aware Loss (SAL). The purpose of SAL is to encourage the policy to learn representations that are aware of the current subtask stage. As a result, the policy should not only understand the overall task goal, but also recognize the current subtask stage from the observation and the language condition.

Let $\mathcal{B} = \{b^0, b^1, \ldots, b^S\}$ denote the subtask boundaries, where $b^0$ and $b^S$ correspond to the start and end of a trajectory, respectively. The timestep $t$ is assigned to the $s$-th subtask stage if
\begin{equation}
    b^{s} \leq t < b^{s+1},
    \quad s \in \{0,1,\ldots,S-1\}.
    \label{subtask_boundaries}
\end{equation}

The ground-truth subtask label is defined as
\begin{equation}
    y_t = s,
    \quad \text{if } b^{s} \leq t < b^{s+1},
    \label{eq:subtask_label}
\end{equation}
In our experiments, we set $S=4$ for all demonstrations. 

We formulate this objective as a subtask classification task. Given the latent representation tokens produced by the MDT encoder, we first aggregate them into a single representation vector $\mathbf{z}_i$. This representation is then passed into a subtask classification head $h_{\mathrm{SAL}}$ to predict the current subtask stage. The predicted subtask probability is computed as
\begin{equation}
    \mathbf{p}_i = \operatorname{softmax}\left(h_{\mathrm{SAL}}(\mathbf{z}_i)\right),
    \label{eq:sal_probability}
\end{equation}
where $\mathbf{p}_i \in \mathbb{R}^{S}$ denotes the predicted probability distribution over $S$ subtask classes. 

We use cross-entropy loss to train the subtask classification objective:
\begin{equation}
    \mathcal{L}_{\mathrm{SAL}}
    =
    -\frac{1}{B}
    \sum_{i=1}^{B}
    \log \mathbf{p}_i(y_i),
    \quad
    y_i \in \{0,1,\ldots,S-1\},
    \label{eq:sal_loss}
\end{equation}
where $B$ is the batch size, $S$ is the number of subtask classes, and $y_i$ is the ground-truth subtask label of the $i$-th training sample. By predicting the label of the current subtask, SAL encourages the encoder to encode stage-level information for policy learning.

The overall training objective of MuGIL is defined as
\begin{equation}
    \mathcal{L}_{\mathrm{MuGIL}}
    =
    \mathcal{L}_{\mathrm{MDT}}
    +
    \lambda \mathcal{L}_{\mathrm{SAL}},
\label{eq:mugil_loss}
\end{equation}
where $\lambda=0.2$ in our experiments. 

\section{Experiments}
\label{chap:exp}

\subsection{Experimental Settings}
\label{sec:experimental_settings}
We evaluate our method on four task suites from the LIBERO benchmark~\cite{LIBERO}. In our experiments, we use 20 demonstrations per task for training. During evaluation, we evaluate each method with 20 rollouts per task. We report the average success rate over all tasks in each suite. 

We use a single NVIDIA GeForce RTX 4080 GPU (CUDA 13.0) with 32 logical CPU threads. In our experiments, all methods are trained with the observation images from the static camera. All input images are normalized using the CLIP image normalization statistics. We apply random shift augmentation to the observation images during training and resize them to $224 \times 224$ pixels. For action prediction, we use the default end-effector action space in all our experiments. All models are trained for 50 epochs, and we used the checkpoint from the final training epoch. The detailed hyperparameters are provided in Table~\ref{tab:hyperparameters}.

\begin{table}
    \centering
    \caption{Hyperparameters of policy training in this work.}
    \label{tab:hyperparameters}
    \scriptsize
    \begin{tabular}{l c | l c}
        \hline
        \textbf{Hyperparameter} & \textbf{Value} & \textbf{Hyperparameter} & \textbf{Value} \\
        \hline
        epochs & 50 & batch size & 64 \\
        \# encoder layers & 4 & \# decoder layers & 6 \\
        attention heads & 8 & action chunk size & 10 \\
        history length & 1 & goal window sampling size & 49 \\
        hidden dimension & 512 & image encoder & ResNet18 \\
        attention dropout & 0.3 & residual dropout & 0.1 \\
        MLP dropout & 0.05 & input dropout & 0.0 \\
        optimizer & AdamW & betas & [0.9, 0.9] \\
        learning rate & 1e-4 & weight decay & 0.05 \\
        other weight decay & 0.05 & trainable parameters & 78.3 M \\
        $\sigma_{\max}$ & 80 & $\sigma_{\min}$ & 0.001 \\
        $\sigma_t$ & 0.5 & time steps & Exponential \\
        sampler & DDIM & language goal encoder & CLIP ViT-B/32 \\
        \hline
    \end{tabular}
\end{table}

\begin{table*}
    \centering
    \caption{Success rates on different LIBERO task suites. 
    The best results are shown in bold, and the second-best results are underlined.}
    \label{tab:libero_results}
    \renewcommand{\arraystretch}{1.2}
    \begin{tabular}{l c c c c c c c}
        \hline
        Method 
        & \begin{tabular}{c} Mixed \\ Language \end{tabular}
        & $\mathcal{L}_{\mathrm{SAL}}$
        & \begin{tabular}{c} LIBERO- \\ Goal \end{tabular}
        & \begin{tabular}{c} LIBERO- \\ Object \end{tabular}
        & \begin{tabular}{c} LIBERO- \\ Spatial \end{tabular}
        & \begin{tabular}{c} LIBERO- \\ 10 \end{tabular}
        & Avg. \\
        \hline
        MDT 
        & 
        & 
        & 63.0 
        & 88.0 
        & 73.5 
        & \underline{48.0} 
        & 68.13 \\
        \hline
        \multirow{2}{*}{MuGIL}
        & \checkmark
        & 
        & \underline{70.0}
        & \textbf{96.0}
        & \textbf{77.5}
        & 46.5
        & \underline{72.50} \\
        & \checkmark
        & \checkmark
        & \textbf{77.5}
        & \underline{94.5}
        & \underline{73.5}
        & \textbf{63.5}
        & \textbf{77.25} \\
        \hline
    \end{tabular}

\end{table*}

\subsection{Performance on the LIBERO benchmark}
\label{sec:performance}
Table~\ref{tab:libero_results} reports the success rates of different methods on the LIBERO benchmark. We compare the baseline MDT policy with our proposed MuGIL framework under two settings: using the mixed language instruction strategy only, and further incorporating the subtask-aware loss $\mathcal{L}_{\mathrm{SAL}}$.

Overall, MuGIL achieves better performance than the baseline MDT. Compared with MDT, applying the mixed language instruction strategy improves the average success rate from 68.13\% to 72.5\%. This shows that training the policy with the language guidance at different levels can provide richer information for policy learning. In particular, the mixed language strategy improves the performance on LIBERO-Goal, LIBERO-Object, and LIBERO-Spatial by 7.0\%, 8.0\%, and 4.0\%, respectively. The results suggest that fine-grained language guidance can help the policy better understand the current manipulation context.

However, on LIBERO-10, the success rate slightly decreases from 48.0\% to 46.5\% when only the mixed language instruction strategy is applied. One reason is that LIBERO-10 contains longer-horizon tasks that require completing multiple subtask stages sequentially. For such tasks, maintaining a stable and consistent task objective throughout the trajectory becomes more important. Although mixed language instructions provide more diverse language information, they may also introduce ambiguity if the policy does not explicitly recognize the current subtask stage. After incorporating $\mathcal{L}_{\mathrm{SAL}}$, the improvement is especially significant on LIBERO-10, where the success rate increases from 46.5\% to 63.5\%. This result suggests that $\mathcal{L}_{\mathrm{SAL}}$ helps the policy capture subtask stage information and improves its subtask awareness. By explicitly encouraging the learned representation to encode the current execution stage, the policy can better handle long-horizon tasks.

Although adding $\mathcal{L}_{\mathrm{SAL}}$ slightly reduces the performance on LIBERO-Object and LIBERO-Spatial compared with using mixed language alone, 
it still achieves the highest average success rate across all suites. These results indicate that combining multi-granularity language guidance with subtask-aware supervision provides a more effective learning signal.

\begin{figure}
    \centering
    \subfloat{\includegraphics[width=0.48\linewidth]{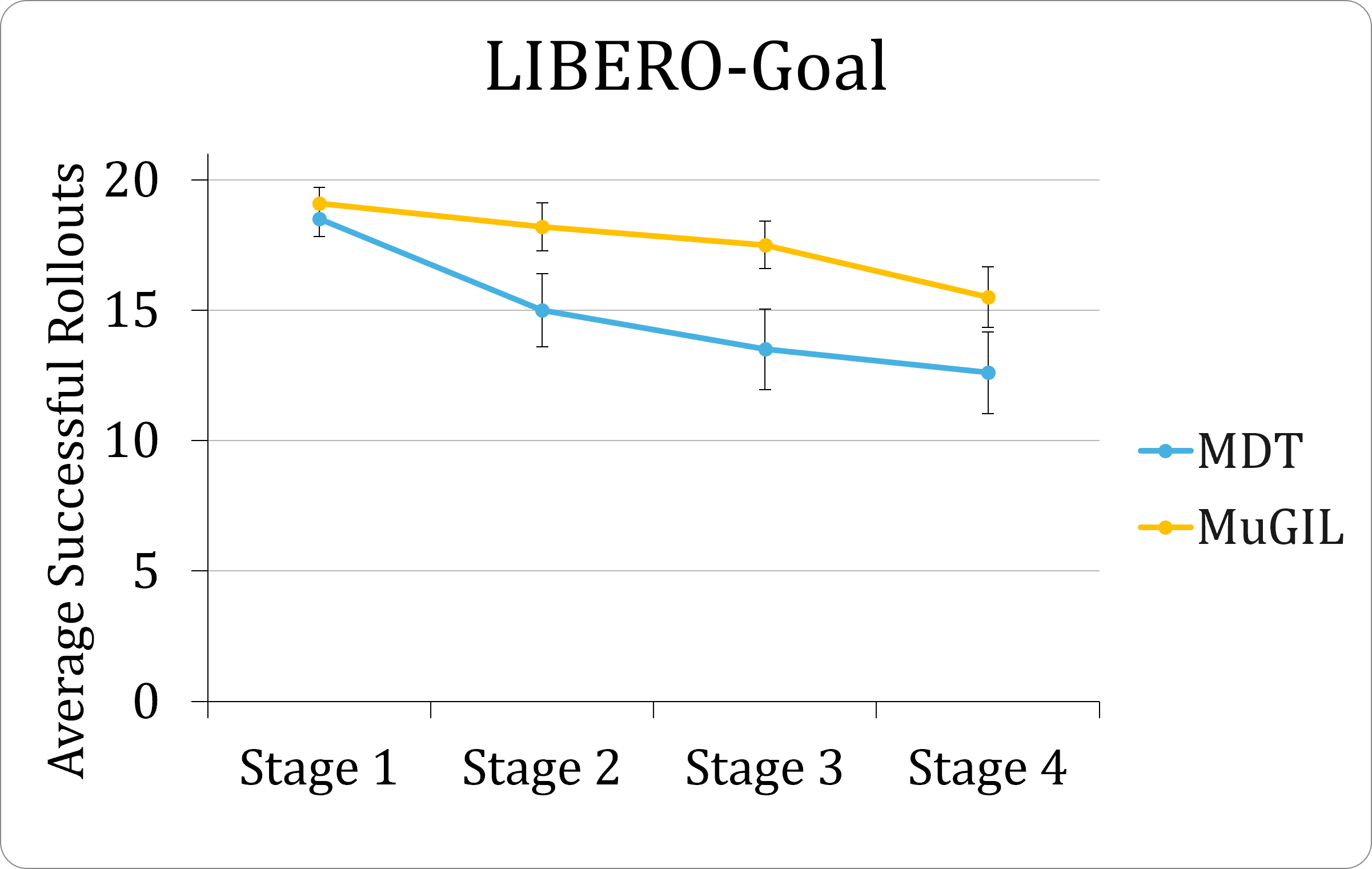}}
    \subfloat{\includegraphics[width=0.48\linewidth]{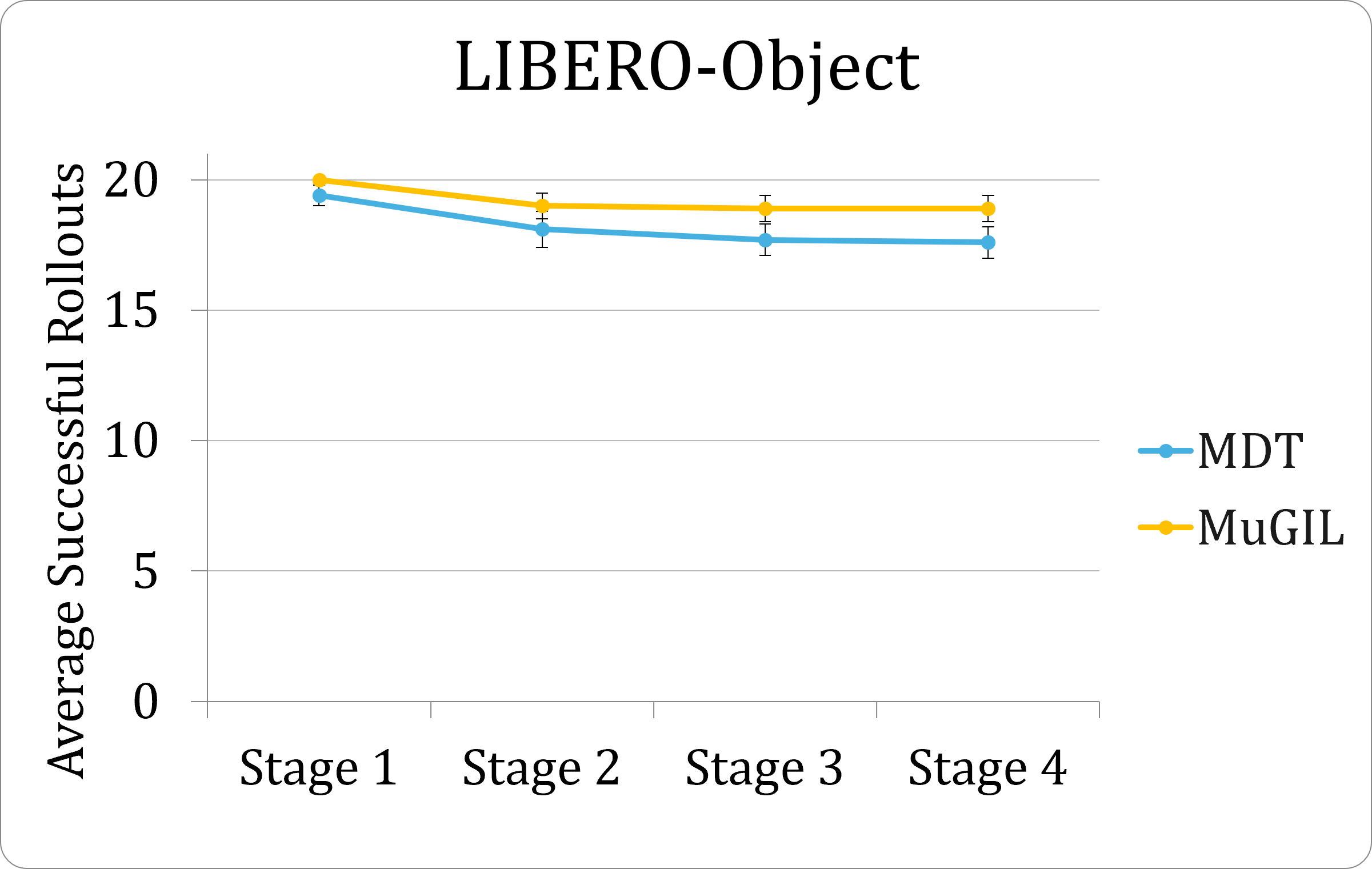}}
    \hfill
    \subfloat{\includegraphics[width=0.48\linewidth]{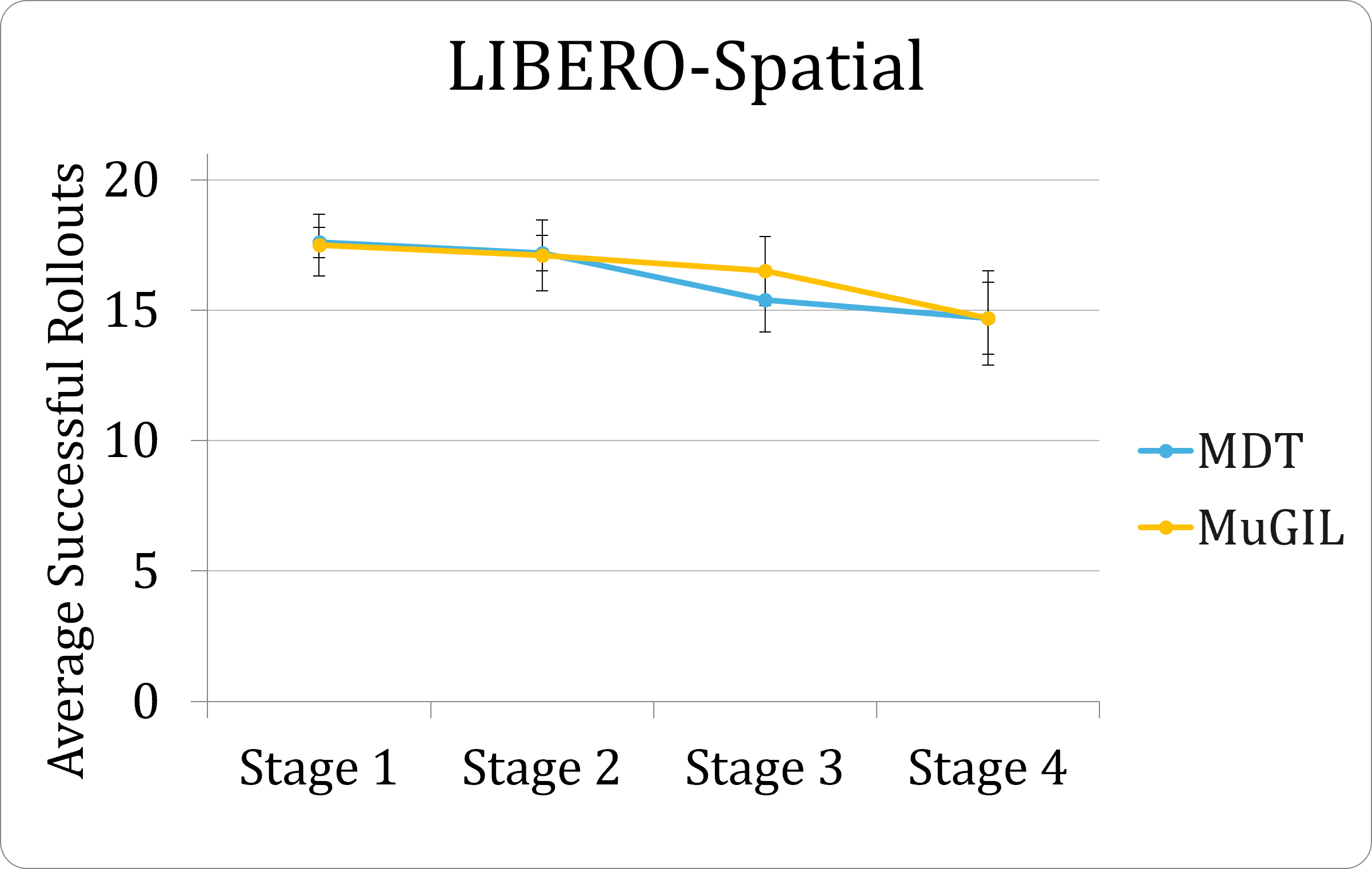}}
    \subfloat{\includegraphics[width=0.48\linewidth]{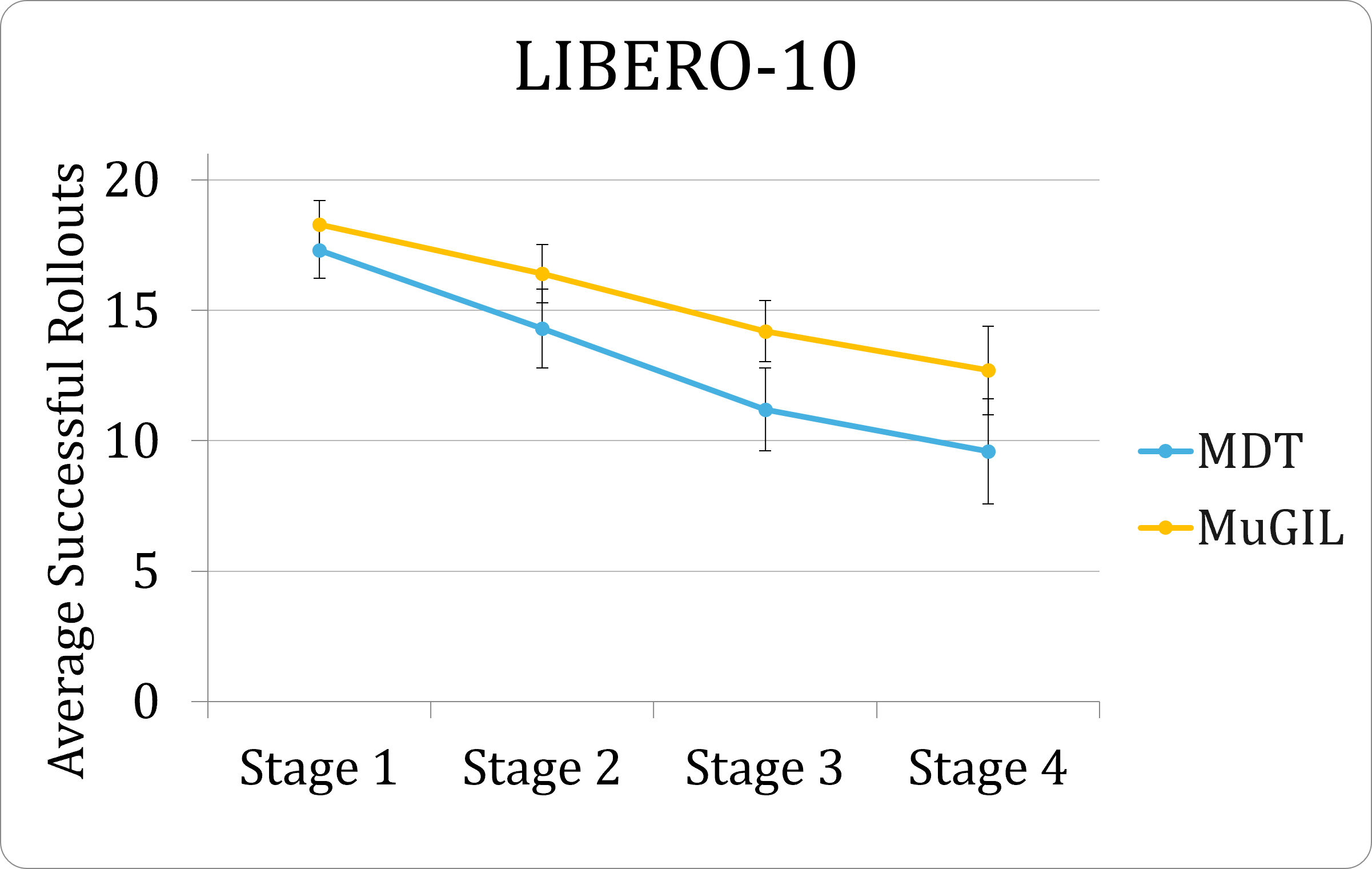}}
    \caption{Stage-wise success counts on different LIBERO task suites.}
    \label{fig:stage_wise}
\end{figure}

\subsection{Ablation Study}
\label{sec:ablation_study}

\subsubsection{Stage-Wise Success Counts}
\label{sec:stage_wise_success_counts}
To further analyze whether MuGIL improves task execution across different subtask stages, we evaluate the stage-wise success counts for each task suite, as shown in Figure~\ref{fig:stage_wise}. Inspired by the stage-wise evaluation in RoboEval~\cite{Roboeval}, we define the stage-wise success count as the number of successful rollouts at each subtask stage, rather than only measuring whether the entire task is completed. This metric provides a more detailed view of how well a policy progresses through different subtask stages.

In Figure~\ref{fig:stage_wise}, the yellow curve represents our proposed MuGIL method, while the blue curve represents the baseline MDT trained only with coarse language instructions. Across the four LIBERO task suites, MuGIL generally achieves higher stage-wise success counts than MDT. This indicates that the proposed multi-granularity language guidance and subtask-aware learning help the policy complete more intermediate subtask stages during execution.

The improvement is especially clear on LIBERO-Goal and LIBERO-10, which suggests that MuGIL is more effective at maintaining task progress when the policy is required to complete multiple subtask stages sequentially. For LIBERO-Object and LIBERO-Spatial, the early-stage success counts remain relatively high for both MDT and MuGIL. Nevertheless, MuGIL still maintains comparable or better performance across later stages.

These results suggest that MuGIL not only improves final task success but also enhances the policy's ability to complete intermediate subtask stages, demonstrating better subtask-level execution capability.

\begin{table*}
    \centering
    \caption{Comparison between semantic and non-semantic fine-grained language.}
    \label{tab:semantic_fine}
    \renewcommand{\arraystretch}{1.2}
    \begin{tabular}{l c c c c c}
        \hline
        Method 
        & \begin{tabular}{c} LIBERO- \\ Goal \end{tabular}
        & \begin{tabular}{c} LIBERO- \\ Object \end{tabular}
        & \begin{tabular}{c} LIBERO- \\ Spatial \end{tabular}
        & \begin{tabular}{c} LIBERO- \\ 10 \end{tabular}
        & Avg \\
        \hline
        MuGIL w/ Semantic Fine
        & \textbf{77.5} 
        & \textbf{94.5} 
        & \textbf{73.5} 
        & \textbf{63.5} 
        & \textbf{77.25} \\
        \hline
        MuGIL w/ Non-Semantic Fine
        & 72.5 
        & 91.0 
        & 65.0 
        & 53.5
        & 70.50 \\
        \hline
    \end{tabular}
\end{table*} 

\begin{table*}
    \centering
    \caption{Different language mode sampling ratios on different LIBERO task suites.}
    \label{tab:language_mode_ratio}
    \renewcommand{\arraystretch}{1.2}
    \begin{tabular}{l ccc ccccc}
        \hline
        \multirow{2}{*}{Method}
        & \multicolumn{3}{c}{Sampling Ratio}
        & \multirow{2}{*}{LIBERO-Goal}
        & \multirow{2}{*}{LIBERO-Object}
        & \multirow{2}{*}{LIBERO-Spatial}
        & \multirow{2}{*}{LIBERO-10}
        & \multirow{2}{*}{Avg.} \\
        & Coarse & Fine & Both
        &  &  &  &  &  \\
        \hline
        \multirow{3}{*}{MuGIL}
        & 0.5 & 0.2 & 0.3 & 77.0 & 94.0 & 68.0 & 53.5 & 73.13 \\
        & 0.6 & 0.1 & 0.3 & \textbf{77.5} & \textbf{94.5} & \textbf{73.5} & \textbf{63.5} & \textbf{77.25} \\
        & 0.7 & 0.0 & 0.3 & 74.0 & 86.0 & 69.0 & 56.5 & 71.38 \\
        \hline
    \end{tabular}
\end{table*}

\subsubsection{Semantic Fine Language Guidance}
\label{sec:semantic_fine_language_guidance}
To further investigate the importance of semantic information in fine-grained language guidance, we compare our proposed method under two settings: semantic fine language and non-semantic fine language. Semantic fine language refers to fine language instructions that contain detailed semantic information about the current subtask, while non-semantic fine language refers to fine language instructions that provide only simple step identifiers without detailed semantic descriptions.

In the original design, the fine language instruction contains meaningful subtask descriptions, such as \enquote{Move the gripper towards the milk} or \enquote{Close the gripper to pick the milk.} These instructions provide detailed semantic information about the current manipulation stage. For comparison, we construct a non-semantic fine language setting. In this setting, the original coarse language instruction remains unchanged, but the fine language instruction is replaced with non-semantic step identifiers, such as \enquote{Task 1. Step 1} and \enquote{Task 1. Step 2.} Similarly, for the both language mode, the subtask description is replaced with the corresponding non-semantic step identifier. This setting preserves the temporal subtask information but removes the semantic meaning of each fine-grained instruction. Therefore, this ablation allows us to examine whether the performance improvement comes from the semantic content of fine language or merely from the additional subtask stage indicator.

Table~\ref{tab:semantic_fine} shows the comparison between semantic and non-semantic fine language guidance. MuGIL with semantic fine language consistently outperforms the non-semantic variant across all LIBERO task suites. The average success rate decreases from 77.25\% to 70.50\% when the semantic fine language is replaced with non-semantic step identifiers. This result indicates that the semantic content of fine-grained language plays an important role in policy learning. 

This suggests that meaningful subtask descriptions provide useful local guidance for distinguishing different manipulation stages. Although the non-semantic setting still provides information about the index of the current subtask, it does not describe what manipulation behavior should be performed at that stage. As a result, the policy receives less informative guidance and achieves lower performance.

\subsubsection{Language Mode Sampling Ratios}
\label{sec:sampling_ratio} 
As described in Section~\ref{sec:mixed_language_instruction_strategy}, one of the three language modes is sampled during training, and the sampling ratio controls how frequently each mode of language guidance is used. To analyze the influence of different language mode sampling ratios, we vary the sampling ratios of the coarse, fine, and both language modes.

Table~\ref{tab:language_mode_ratio} shows the results under different sampling ratios. The setting with the ratio $p_c:p_f:p_b=0.6:0.1:0.3$ achieves the best overall performance. This suggests that a balanced combination of coarse, fine, and both language instructions provides the most effective supervision. When the sampling ratio of fine language instruction is removed, as in the ratio $0.7 : 0.0 : 0.3$, the performance decreases noticeably. This indicates that explicitly training the policy with fine-only language conditions is important, as it helps the policy learn how to follow fine language instructions directly. On the other hand, reducing the sampling ratio of coarse language instruction, as in the ratio $0.5:0.2:0.3$, also leads to lower performance across all task suites. This result implies that coarse language instruction remains important for providing task-level guidance, as reducing its sampling ratio weakens the supervision of the overall task objective. This makes it harder for the policy to maintain a consistent task objective during execution. Overall, the proper sampling ratio $0.6:0.1:0.3$ provides a better balance between global task guidance and local subtask guidance. 

\section{CONCLUSION}
\label{chap:conclusion}
We propose MuGIL, a multi-granularity language guidance framework for language-guided imitation learning. Instead of relying only on a single language instruction, MuGIL decomposes it into multiple finer language instructions that describe different subtasks. By providing language guidance at different levels, the policy can receive both global task-level guidance and local subtask-level guidance during training. To effectively incorporate different forms of language guidance, we introduce a mixed language instruction strategy, which allows the policy to learn from coarse, fine, and both language instructions. In addition, we design a subtask-aware loss to encourage the policy to distinguish different subtasks, thereby enhancing its awareness of task progress.

We evaluate MuGIL on the LIBERO benchmark across multiple task suites. The experimental results show that MuGIL improves policy performance compared with the baseline MDT. In particular, the mixed language instruction strategy provides more informative language supervision, while the proposed SAL further improves performance on more challenging long-horizon tasks. The ablation studies further show that MuGIL can maintain more successful rollouts across different subtasks, 
and that the semantic content of fine language instructions plays an important role in policy learning.

In the future, we can investigate how to use fine language instructions not only as training supervision but also as an explicit mechanism for subtask switching during execution. Instead of relying on predefined temporal segments, the policy can automatically determine whether the current subtask has been completed and switch to the next subtask accordingly. Such an online subtask transition mechanism may further improve the flexibility and robustness of long-horizon manipulation policies.

\textbf{Acknowledgement. }
This work was funded in part by the National Science and Technology Council, Taiwan, under grants 115-2622-8-006-015, 114-2622-E-006-028, 114-2221-E-006-047-MY3, 115-2425-H-006-005, 115-2218-E-006-024, and 114-2634-F-006-002.

\bibliographystyle{IEEEtran}
\bibliography{references}

\end{document}